\documentclass[acmsmall]{acmart}
\usepackage{booktabs}
\usepackage{amsmath}
\usepackage{amsthm}
\usepackage{color}
\usepackage{booktabs}
\usepackage{algorithm}
\usepackage{algorithmic}
\usepackage{graphicx}
\AtBeginDocument{%
  \providecommand\BibTeX{{%
    \normalfont B\kern-0.5em{\scshape i\kern-0.25em b}\kern-0.8em\TeX}}}

\setcopyright{acmcopyright}
\copyrightyear{2018}
\acmYear{2018}
\acmDOI{10.1145/1122445.1122456}

\acmJournal{JACM}
\acmVolume{37}
\acmNumber{4}
\acmArticle{111}
\acmMonth{8}

\begin{document}

\title[MG-GCN: Fast and Effective Learning with Mix-grained Aggregators]{MG-GCN: Fast and Effective Learning with Mix-grained Aggregators for Training Large Graph Convolutional Networks}

\author{Tao Huang}
\email{huangt97@mail2.sysu.edu.cn}
\affiliation{%
  \institution{School of Data and Computer Science, Sun Yat-sen University}
  \city{Guangzhou}
  \state{Guangdong}
  \country{China}
}
\author{Yihan Zhang}
\email{zhangyihan76@gmail.com}
\affiliation{%
  \institution{School of Data and Computer Science, Sun Yat-sen University}
  \city{Guangzhou}
  \state{Guangdong}
  \country{China}
}
\author{Jiajing Wu}
\authornote{Contact Author.}
\email{wujiajing@mail.sysu.edu.cn}
\affiliation{%
  \institution{School of Data and Computer Science, Sun Yat-sen University}
  \city{Guangzhou}
  \state{Guangdong}
  \country{China}
}
\author{Junyuan Fang}
\email{fangjy26@mail2.sysu.edu.cn}
\affiliation{%
  \institution{School of Data and Computer Science, Sun Yat-sen University}
  \city{Guangzhou}
  \state{Guangdong}
  \country{China}
}
\author{Zibin Zheng}
\email{zhzibin@mail.sysu.edu.cn}
\affiliation{%
  \institution{School of Data and Computer Science, Sun Yat-sen University}
  \city{Guangzhou}
  \state{Guangdong}
  \country{China}
}
\renewcommand{\shortauthors}{Tao and Yihan, et al.}

\begin{abstract}
   Graph convolutional networks (GCNs) have been employed as a kind of significant tool on many graph-based applications recently. Inspired by convolutional neural networks (CNNs), GCNs generate the embeddings of nodes by aggregating the information of their neighbors layer by layer. 
However, the high computational and memory cost of GCNs due to the recursive neighborhood expansion across GCN layers makes it infeasible for training on large graphs. To tackle this issue, several sampling methods during the process of information aggregation have been proposed to train GCNs in a mini-batch Stochastic Gradient Descent (SGD) manner. Nevertheless, these sampling strategies sometimes bring concerns about insufficient information collection, 
%such as the sparse connection problem in FastGCN, 
which may hinder the learning performance in terms of accuracy and convergence. 
To tackle the dilemma between accuracy and efficiency, we propose to use aggregators with different granularities to gather neighborhood information in  different layers. Then, a degree-based sampling strategy, which avoids the exponential complexity, is constructed for sampling a fixed number of nodes. Combining the above two mechanisms, the proposed model, named \underline{M}ix-\underline{g}rained GCN (MG-GCN) achieves state-of-the-art performance in terms of accuracy, training speed, convergence speed, and memory cost through a comprehensive set of experiments on four commonly used benchmark datasets and a new Ethereum dataset.
\end{abstract}




\maketitle

\section{Introduction}
Graphs, which consist of a set of objects and relationships, have become increasingly popular owing to their expressive power for non-Euclidean models. Existing research studies and analyses about non-Euclidean data structures usually focus on some typical tasks such as node classification~\cite{li2016discriminative}, link prediction~\cite{wei2017cross}, and clustering~\cite{wang2017community}. To deal with these tasks on large graphs, a series of machine learning methods, such as DeepWalk~\cite{perozzi2014deepwalk}, node2vec~\cite{grover2016node2vec}, and Graph2Vec~\cite{mlg2017_21}, have been proposed to represent nodes, edges  or sub-graphs in low-dimensional vectors. However, most of  these graph representation methods suffer from the increasing number of parameters with the number of nodes and the lack of generalization to new nodes or new graphs.

Inspired by the great success of convolutional neural networks (CNNs)~\cite{rawat2017deep} on pattern recognition and data mining, the notation of \textit{convolution} has been re-defined for graph data and the corresponding convolution architecture is referred to as graph convolutional network (GCN)~\cite{wu2019comprehensive}. GCNs generate the embedding of a node by aggregating the embeddings of its neighbors through linear transformations and nonlinear activations \cite{nwankpa2018activation} layer by layer. Similar to CNNs, the parameters of GCNs are shared among all nodes and do not scale with the number of nodes. Moreover, the trained model could be easily generalized to an unseen node.

%Different from the limited size of kernels in CNNs, the convolutional operator in GCNs can be classified into two parts, propagation and aggregation. The propagation part is used to collect neighbors layer by layer, the information of which are gathered to derive the hidden representation through aggregation parts.

Different from the limited size of kernels in CNNs, the convolutional operator in GCNs could be expensive because it depends on the coupling relationship of the graph data and the number of final involved nodes could be extremely large. Furthermore, one of the most famous GCN models~\cite{kipf2017semi} is trained under a full-batch manner, which requires calculating the loss term for all objects (e.g., classification loss on nodes or link prediction loss on edges) at once. The above two mechanisms bring high computational and memory costs when training GCNs in large graphs.

To alleviate these issues, several sampling methods have been proposed to train GCNs in a mini-batch SGD  manner~\cite{DBLP:journals/siamrev/BottouCN18,ruder2016overview} . 
Among them, GraphSAGE~\cite{hamilton2017inductive} proposed to sample a fixed size of neighbors uniformly for each node recursively. This kind of sampling reduces the costs to some extent, but the time and memory complexity still grows exponentially with the number of GCN layers. In addition, uniform sampling on first-order neighbors also raises concerns about insufficient information collection. Here we conduct a simple experiment to demonstrate the importance of neighbors from different orders. As shown in Table ~\ref{First-order important}, the results are measured by the accuracy of node classification. Though the time and memory costs of the full batch GCN grow exponentially with the number of layers, the prediction performance does not necessarily increase. Instead, the 1-layer model incorporating information from first-order neighbors can derive a relatively high accuracy. This phenomenon indicates that the first-order neighbors, part of which are abandoned in GraphSAGE, sometimes play a more significant role than the higher-order neighbors.
\begin{table*}[tb]
\caption{Results of 1-layer, 2-layer, and 3-layer full-batch GCN on Cora, Citeseer, and Pubmed.}
	\label{First-order important}
	\centering
	
	\begin{tabular}{p{3cm}p{3cm}p{3cm}p{2cm}}  
		\toprule
		Dataset  &  1-layer & 2-layer & 3-layer \\
		\midrule
		Cora       & 0.849  & \textbf{0.859} &  0.837   \\
		
		Citeseer    & \textbf{0.768}  & \textbf{0.768}  &  0.751  \\
		Pubmed   & \textbf{0.88}  & 0.863   &  0.851 \\
		
		\bottomrule
	\end{tabular}
\end{table*}
Along the lines of GraphSAGE, VR-GCN~\cite{DBLP:conf/icml/ChenZS18} proposed to use a variance reduction technique to reduce the sampling complexity. However, as pointed out in~\cite{DBLP:conf/kdd/ChiangLSLBH19}, VR-GCN requires storing all the historical embeddings of all nodes in memory, which results in large memory costs when the number of nodes grows. 

Another method, named FastGCN~\cite{chen2018fastgcn}, introduced a layer-wise sampling method to improve training efficiency by orders of magnitude. Nevertheless, a recent study~\cite{NIPS2019_9303} showed that FastGCN suffers from the problem of sparse connection, which can be considered as a kind of insufficient information collection problem caused by independent sampling probability for each layer. Such a problem would bring negative effects on both accuracy and convergence. To mitigate this problem, AS-GCN~\cite{huang2018adaptive} proposed to sample the lower layer conditioned on the top one by an adaptive sampler, which enjoys higher accuracy and convergence but sacrifices the training speed. Different from the afore-mentioned sampling methods, SGC~\cite{Wu2019SimplifyingGC} reduced 
the complexity and improved the training efficiency by removing nonlinearities and collapsing weight matrices between consecutive layers. Yet at the same time, under SGC, the simplification of the layer structure leads to the decline of accuracy.
To sum up, all these algorithms still cannot solve the dilemma of keeping accuracy and convergence while reducing computational and memory cost.

%GraphSAGE and FastGCN we have discussed above while not suffering from the unbearable time and memory complexity
%in consideration of both accuracy and efficiency 

Inspired by the significance of first-order neighbors in information aggregation, we propose a GCN model with mix-grained aggregators that can aggregate first-order neighborhood information in a fine-grained way and gather high-order information in a more coarse-grained manner. Specifically, when the fine-grained aggregator is applied, all one-hop neighbors of a node will be aggregated whether they are sampled or not. On the contrary, the coarse-grained aggregator only aggregates the information of sampled nodes. By combining both kinds of aggregators, GCNs can achieve a win-win situation in both accuracy and efficiency. Besides, a new degree-based sampling method with a fixed sample size is proposed to sample neighbors while avoiding exponential complexity.

Specifically, the main contributions of this work are summarized as follows.
\begin{itemize}
	\item We propose a new GCN-based model, named \underline{M}ix-\underline{g}rained GCN (MG-GCN) with two kinds of aggregators with different granularities, namely, fine-grained aggregators and coarse-grained aggregators, helping avoid the insufficient information collection problem and reduce the computational and memory cost simultaneously.
	\item We introduce a new degree-based sampling strategy to sample neighbors from different layers. By setting a fixed sample size which is independent with the number of layers, this strategy can explore neighboring nodes with larger degrees while avoiding the exponential increment of sampled neighbors.
	\item Experiments on four node classification benchmarks and a new collected Ethereum dataset, demonstrate the proposed model can achieve state-of-the-art performance in terms of accuracy, memory cost, training and convergence speed. 
\end{itemize}
\newcommand{\tabincell}[2]{\begin{tabular}{@{}#1@{}}#2\end{tabular}}

\section{Preliminary}
\begin{table*}[tb]
\caption{Notations in this paper.}
	\label{Notations}
	\centering
	\begin{tabular}{p{2cm}p{9cm}}  
		\toprule
		$G=(V,E)$  &  A input graph with node set $V$ and edge set $E$ \\
		\midrule
		$v, u$  &  Nodes in graph \\
		\midrule
		$B$  &  A batch of nodes \\
		\midrule
		$W^{i}$  &  The parameter matrix in $i$-th layer of GCN \\
		\midrule
		$b, d, L$  &  The batch size, averaged degree and total number of layers\\ 
		\midrule
		$\hat{A}$  &  A normalization of the graph adjacency matrix \\
		\midrule
		$\sigma(\cdot)$  &  The nonlinear activation function in GCN \\
		\midrule
		$H^{l}$   &  The hidden representation of all nodes in layer $l$ \\		
		\midrule
		$H_{v}^{l}$   &  The hidden representation of node $v$ in layer $l$ \\
		\midrule
		$loss(\cdot, \cdot)$   & A loss function such as cross entropy \\
		\midrule
		$y_{v}$   &  The label of node $v$ \\
		\midrule
		$N(v)$   &  The set of one-hop neighbor nodes of node $v$ \\
		\midrule
		$p(v)$   &  The sampling probability of node $v$ \\
		\midrule
		$S(v)$   &  The set of nodes sampled from node $v$\\
		\bottomrule
	\end{tabular}
\end{table*}
In this section, we will introduce the basic architecture of the GCN model and some of its variants. Notations used in this paper are described in Table \ref{Notations}.

In the GCN model proposed in ~\cite{kipf2017semi}, the convolution operation in every layer is given as:
\begin{align}
	H^{l+1} = \sigma(\hat{A}H^{l}W^{l+1}),
\end{align}%
and the hidden representation of the last layer will be utilized for downstream tasks. The parameter matrix $W^{l}$ of neural networks can be learned by minimizing the loss function, i.e.,
\begin{align}
	Loss_{full} = \dfrac{1}{\left|V\right|}\sum_{v \in V}loss(H_{v}^{L}, y_{v}).
\end{align}%
This full batch learning results in high computation and memory costs when applying to large graphs. Intuitively, we can conduct the learning process in a mini-batch style:
\begin{align}
	Loss_{mini} = \dfrac{1}{\left|B\right|}\sum_{v \in B}loss(H_{v}^{L}, y_{v}).
\end{align}%

However, the calculation for $H_{v}^{L}$ depends on neighbors of $v$ , which exponential propagates layer by layer. Therefore, several sampling-based methods, such as GraphSAGE and FastGCN, have been proposed to reduce propagation costs. The convolution operator in these methods only consider on the neighbor nodes which have been sampled in each layer, i.e.,
\begin{align}
	H_{v}^{l+1} = \sigma(f(H_{v}^{l},\{H_{u}^{l}, \forall u\in S(v) \cap N(v)\})\cdot W^{l+1}),
\end{align}%
where $f(\cdot, \cdot)$ represents a specific aggregator, which could be Mean, LSTM~\cite{hochreiter1997long} or Pooling operator.

Particularly, GraphSAGE samples nodes from neighbors uniformly, and the sampling probability of a neighbor node $v$ can be expressed as:
\begin{align}
	p(v)=
	\begin{cases}
		\dfrac{s}{|N(u)|}, & v\in N(u);\\
		0, & \text{otherwise,}
	\end{cases}
\end{align}%
where $u$ is a node that has been sampled and $s$ is the sample size.
As for FastGCN, the sampling possibility of each node from the same layer is identical, which can be expressed as:
\begin{align}
	p(v)=\dfrac{||\hat{A}(:, v)||^2}{\sum_{u\in V}||\hat{A}(:, u)||^2},
\end{align}%
where $p(v)$ is totally independent with its coupling relationship.

Although GraphSAGE and FastGCN can obviousely speed up the training process, they both suffer from the problem of insufficient sampling. For GraphSAGE, this problem is relatively slighter because its sampling strategy is neighbor dependent at least. Nevertheless, the number of involved nodes in GraphSAGE still exponential grows with the number of layers. For FastGCN, insufficient sampling might evolve into the problem of sparse connection due to the layer-independent sampling strategy, which cannot ensure the connectivity between layers.

%Removing redundant activate functions can also speed up GCN, as well as compressing weight matrix. The convolution operation in each SGC layer is given as:
On the other hand, SGC~\cite{Wu2019SimplifyingGC} simplifies and accelarates the training process by removing redundant activate functions and compressing the weight matrix. The convolution operation in each SGC layer is given as:
\begin{align}
	H^{l+1} = \hat{A}H^{l}W^{l+1},
\end{align}%
where the weight matrix $W^{l}$ in each layer can be regarded as a single matrix $W$. 
%This simplified method reduces the number of parameters in the model and improves efficiency, but the reduction of nonlinearities also dramatically reduces the accuracy.

Although this simplified method reduces the number of parameters to improve computational efficiency, the lack of nonlinearity dramatically decreases the accuracy.

\section{Proposed Model}
\label{sec:model}
In this section, we will introduce the base architecture of our model with the motivation behind. We first introduce the concept of fine-grained and coarse-grained aggregators, and then explain how these two kinds of aggregators work together in our model. After that, we introduce our sampling method and a modified coarse-grained aggregator with skip connections.

\subsection{Aggregator}
Different from the previous work that focuses on modifying the sampling strategies while applying the same aggregators in different layers, we employ aggregators with different granularities in different layers. Specifically, when it comes to the low-order layers (e.g., first layer), we aggregate the information of all the one-hop neighbors. For higher-order layers, the aggregator only gather the information of sampled nodes.

\begin{figure}[tb]
	\centering
	\includegraphics[width=0.8\textwidth]{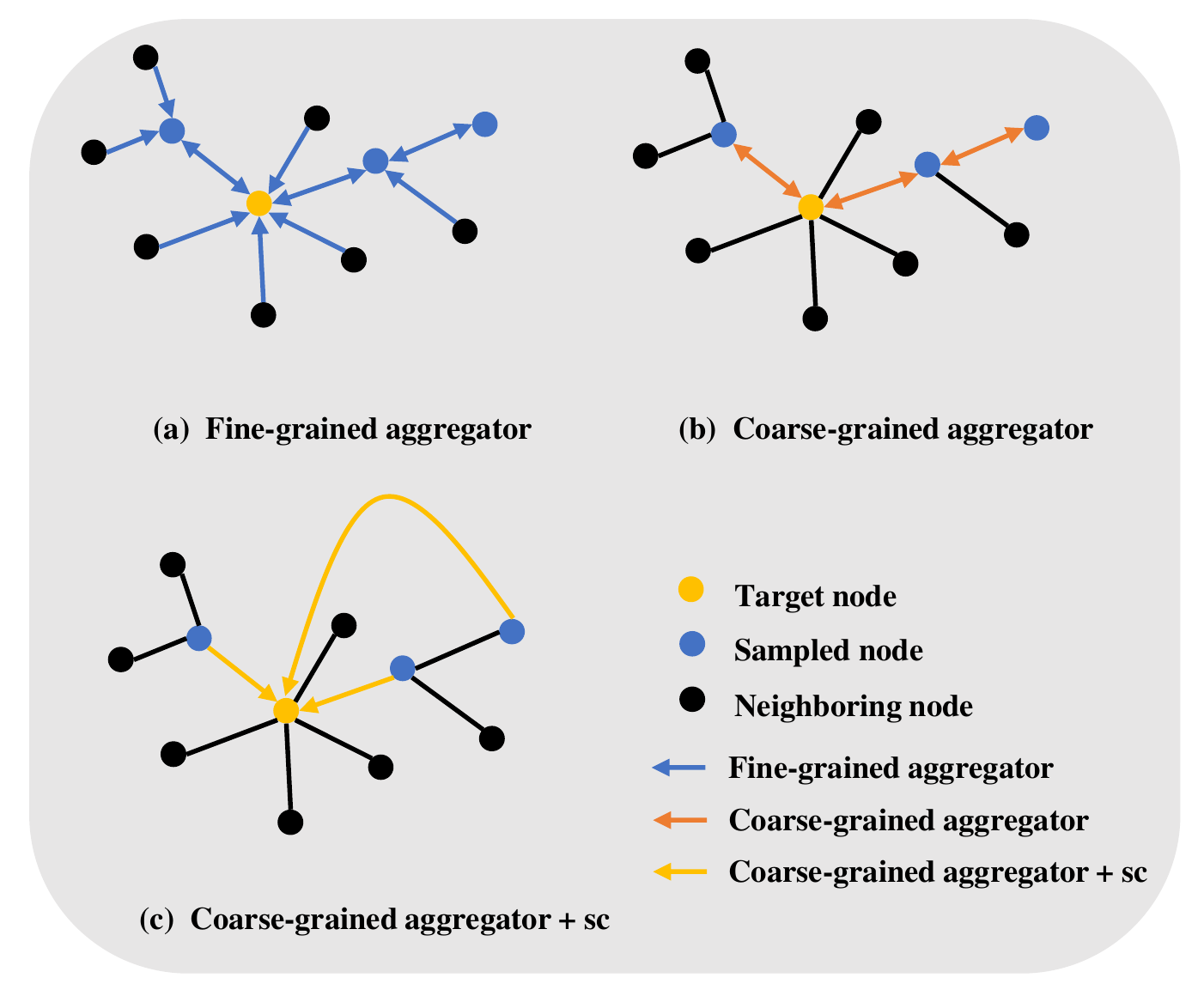}  
	\caption{Visual illustration of different aggregators in MG-GCN. }
	\label{Aggregators}
\end{figure}
The two kinds of aggregators are defined as follows.
\begin{itemize}
	\item \textbf{Fine-grained aggregator:} The fine-grained aggregator, which is shown in Fig. \ref{Aggregators}a, is used to gather the information of all neighbors in low-order layers. The motivation behind the design of this aggregator is the significance of the low-order neighbors we have discussed in Section 1. Furthermore, previous researches on FastGCN also show that the lack of neighbor nodes leads to poor convergence performance. The fine-grained aggregator can be expressed as:
	\begin{align}
		H_{v}^{l+1} = \sigma(f(H_{v}^{l},\{H_{u}^{l}, \forall u\in N(v)\})\cdot W^{l+1}).
	\end{align}%
	\item \textbf{Coarse-grained aggregator:} The coarse-grained aggregator, which is shown in Fig. \ref{Aggregators}b, is the same as the aggregators in previous sampling-based algorithms and it will be applied in higher-order layers. This aggregator will only concentrate on the nodes we have sampled to reduce the computation and time complexity. The coarse-grained aggregator can be formally expressed as:
	\begin{align}
		H_{v}^{l+1} = \sigma(f(H_{v}^{l},\{H_{u}^{l}, \forall u\in S(v) \cap N(v)\})\cdot W^{l+1}),
	\end{align}%
	which is the same as Equation (4).
\end{itemize}

As for $f(\cdot,\cdot)$, we employ a mean-aggregator with a concatenation operation  proposed in~\cite{hamilton2017inductive}, which can be expressed as:
\begin{align}
	f(H_{v}^{l}, \cdot) = CONCAT(H_{v}^{l}, Mean(\cdot)), 
\end{align}%
where $\cdot$ could be $N(v)$ or $S(v) \cap N(v)$. $Mean(\cdot)$ will simply calculate the element-wise mean of the input set and $CONCAT(\cdot, \cdot)$ will concatenate the current representation of node $v$, $H_{v}^{l}$, with the averaged neighborhood vector.

By applying fine-grained aggregators in lower-order layers and coarse-grained aggregators in higher-order layers, our model can balance efficiency and accuracy.
Considering the relatively high computational and memory cost of fine-grained aggregators, in this work, we only apply the fine-grained aggregator in the first-layer. 

\subsection{Sampling Strategy}
As mentioned in Sections 1 and 2, GraphSAGE still suffers from the exponential complexity while FastGCN has the sparse connection problems. To tackle these issues, we propose a new neighbor-dependent sampling strategy, which also avoids the exponential complexity with the number of layers.

To explain our sampling strategy clearly, we maintain a set of candidate nodes for the sampler in every sampling step and the sampler only sample nodes from the candidate nodes. The core procedure of our sampling strategy is how to update this set of candidate nodes.

\begin{algorithm}[tb]
	\caption{Sampling Strategy}
	\begin{flushleft}
    \textbf{Input}: Target node $v$\\
    \textbf{Parameter}: Number of sampling nodes $M$\\
    \textbf{Output}: Set of nodes sampled $S(v)$
    \end{flushleft}
	\begin{algorithmic}[1] %[1] enables line numbers
		\STATE Let $candi_0 = N(v)$.
		\STATE Let $S(v) = \{v\}$.
		\FOR {each $k \in [1,M]$}
		\STATE $u = sample(candi_{k-1} - S(v))$.
		\STATE $candi_k = candi_{k-1} \cup N(u)$.
		\STATE $S(v) = S(v) \cup {u}$.
		\ENDFOR
		\STATE \textbf{return} $S(v)$
	\end{algorithmic}
	\label{Sampling Strategy}
\end{algorithm}
Roughly, our sampling strategy can be seen as a procedure of extending the set of candidate nodes and the whole process is shown in Algorithm \ref{Sampling Strategy}. 
Particularly, given a target node $v$, the candidate nodes are set as the one-hop neighbors of $v$ initially. When a new node $u$ is sampled, the set of candidate nodes will be extended by adding one-hop neighbors of $u$ for the next sampling. By repeating the above process for $M$ times, we can obtain the set of $M$ sampled nodes. Besides, for each time of sampling, we ignore the nodes we have sampled to avoid redundant information.

Given a set of candidate nodes, we sample nodes according to the degree of the node. Formally, we define a probability mass function for each node $v$ from the set of the candidate nodes:
\begin{align}
	p_k(v) = \frac{degree(v)}{\sum_{u\in candi_{k-1}} degree(u)}, 
\end{align}%
where $degree(\cdot)$ returns the degree of the input node and $candi_{k}$ denotes the set of candidate nodes at the $k$-round of sampling. This sampling strategy can cooperate with a fine-grained aggregator to gather the nodes with more neighbors.

Under the proposed sampling strategy, the number of sampled nodes is a fixed number $M$ and is irrelevant to the number of layers. That is to say, our method could avoid the exponential complexity problem in GraphSAGE. Besides, the neighbor-based sampler guarantees the connectivity of sampled nodes. Furthermore, our sampling method can explore both depth and breadth information depending on the degree of neighbors.
\subsection{Skip Connection}
However, the irrelevance between the sampling strategy and the number of layers also brings the problem in training GCNs because we can not control the depth of exploration.
\begin{figure}[tb]
	\centering
	\includegraphics[width=0.7\textwidth]{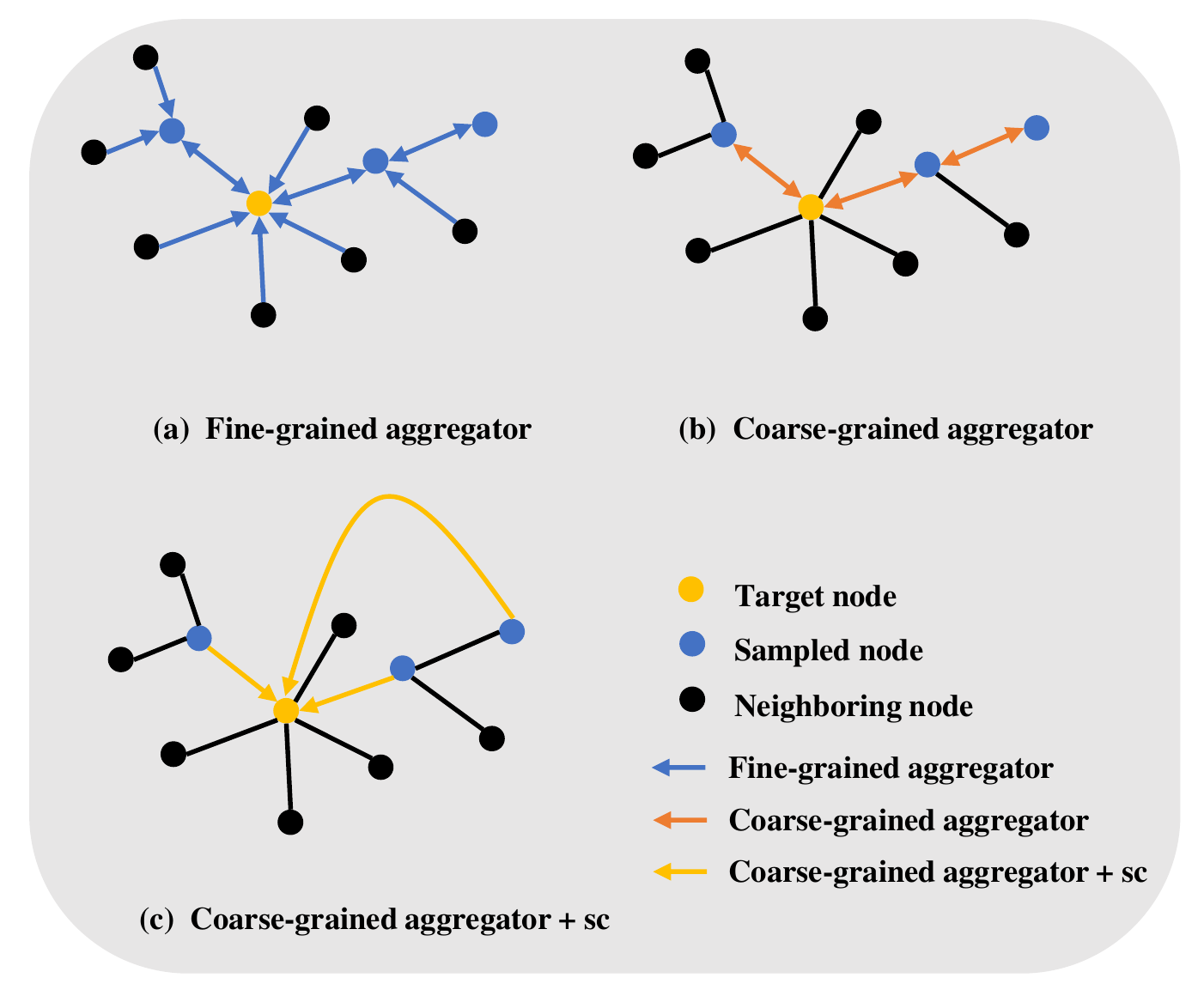}  
	\caption{Visual illustration of sampling nodes.}
	\label{sampling_drawback}
\end{figure}
For example, given a 2-layer MG-GCN with a sample size $M=4$, it is possible for our sampling strategy to sample nodes which are three-hop or four-hop neighbors of the target node just as shown in Fig. \ref{sampling_drawback}. The numbers on nodes means their orders of being sampled. Clearly, in a 2-layer traditional GCN, the fourth sampled node in this figure, which is the 3-hop neighbor of the target node, is useless. The 2-layer GCN can only aggregate the neighbor nodes within two hops, which means the higher-order sampled neighbors are abandoned. 

To tackle this problem, in this work, we proposed a modified coarse-grained aggregator with a skip connection mechanism. As shown in Fig.  \ref{Aggregators}c, skip connections will create direct links between the target node and sampled nodes, and aggregate the higher-order neighbors directly. Formally, this modified coarse-grained aggregator can be expressed as :
\begin{align}
	H_{v}^{l+1} = \sigma(f(H_{v}^{l},\{H_{u}^{l}, \forall u\in S(v)\})\cdot W^{l}),
\end{align}%

To simplify our model, the coarse-grained aggregator with skip connections is only applied in the last layer.

\subsection{Discussion}
To sum up, our model aggregates all one-hop neighbors to generate embeddings for the sampled nodes using fine-grained aggregators in the first layer. Then coarse-grained aggregators are applied in middle layers to reduce complexity. Finally, a modified coarse-grained aggregator with skip connections is utilized in the last layer to aggregate higher-order neighbors directly.
Besides, a sampling method is introduced to explore neighbors with large degrees and it is irrelevant to the number of layers.

Unlike GraphSAGE, the nodes involved in our model does not grow exponentially with the number of layers. In addition, the fully aggregated one-hop neighbors can also improve the accuracy and convergence of the algorithm. 
Moreover, by incorporating both the fine-grained aggregator and the neighbor dependent sampling strategy, the proposed method can avoid the problem of sparse connection which hinder the performance of the FastGCN.

\section{Experiments}

\subsection{Experiment Setups}

\subsubsection{Standard Datasets.} 
\begin{table*}[tb]
	\centering
	\caption{Dataset Statistics.}
	\label{Dataset}
	\begin{tabular}{cccccc}  
		\toprule
		Dataset  &  Nodes (Labeled Nodes) & Edges & Classes & Features& Training/Validation/Test\\
		\midrule
		Cora       & 2,708  & 5,429 &  7 & 1,433& 1,208 / 500 / 1,000  \\
		Citeseer    & 3,327  & 4,732  &  6 & 3,703& 1,827 / 500 / 1,000 \\
		Pubmed   & 1,9717  & 44,338   &  3 & 500& 18,217 / 500 / 1,000\\
		Reddit   & 232,965  & 11,606,919  & 41 &  602& 152,410 / 23,699 / 55,334 \\
	    Ethereum   & 1,402,220 (816) & 2,815,028  & 6 &  13& 572 / 82 / 162 \\
		\bottomrule
	\end{tabular}
\end{table*}
We firstly validate our method on four commonly used benchmark datasets, including three citation network datasets-Cora, Citeseer, Pubmed and a post network-Reddit. We focus on the task of node classification in these datasets. Specifically, our model needs to categorize academic papers in citation network datasets and infer the correct community that different posts belong to in Reddit. According to the statistics summarized in Table \ref{Dataset}, the size of these networks varies from small to large. 
%The number of nodes is $\mathcal{O} (10^3)$ in Cora and Citeseer while $\mathcal{O} (10^4)$ and $\mathcal{O} (10^5)$ in Pubmed and Reddit. 
Besides, we split the training/validation/test sets of these datasets following the scenario in FastGCN, which is more proper for supervised learning.

\begin{table*}[tb]
	\centering
	\caption{Details of the node labels in Ethereum dataset.}
	\label{Ethereum label}
	\begin{tabular}{p{4 cm}p{7cm}}  
		\toprule
		Labels  &  Description \\
		\midrule
		ICO       & ICO wallet used for crowdsale/presale   \\
		Converter    & Automatic token conversion wallet \\
		Exchange   & Places to buy and sell cryptocurrencies  \\
		Mining   & Accounts of mining pools  \\
		Gambling    & Popular gambling contract accounts  \\
		Phish   & Phishing scam accounts  \\
		\bottomrule
	\end{tabular}
\end{table*}
\subsubsection{Ethereum Dataset.} 
To examine the effectiveness of our method in networks of other fields, we create a new Ethereum dataset. Ethereum is an open source public blockchain platform with smart contract capabilities. We obtain 816 accounts with ground-truth node labels from Etherscan\cite{Etherscan}. Then, for each account, we collect its 1-hop neighborhood and randomly select part of 2-hop neighborhood to construct a graph due to the limitation of memory size. 
%Next, due to the limitation of memory size, 
%we randomly collect 10 accounts from these 1-hop neighbors (if the number of 1-hop neighbors is less than 10, all of them are selected), and collect all the 1-hop neighborhood of these selected accounts as the 2-hop neighborhood of the labeled accounts. So far, we can construct a graph through all these collected accounts.
%Finally, we construct a graph through all these collected accounts. 

All these accounts and transaction data were collected through the APIs provided by Etherscan.io. We also filter out the useless transactions by removing all the transactions whose address is ambiguous. The final dataset is a network with 1.4 million nodes and 2.81 million edges (see Table \ref{Dataset} for details). For each node, 13 node features (e.g. balance, in degree, out degree, transaction frequency, etc.) are extracted from the transaction records.
We also focus on the task of node classification on Ethereum dataset as there are 6 types of ground-truth labels (see Table \ref{Ethereum label} for details) among the
training data. Besides, the training:validation:test split of our Ethereum dataset is 7:1:2.

\subsubsection{Baselines.}
We compare the performance of MG-GCN with baselines including GCN \cite{kipf2017semi}, GraphSAGE \cite{hamilton2017inductive}, FastGCN \cite{chen2018fastgcn} and SGC\cite{Wu2019SimplifyingGC}. All baselines and our model are conducted in a 2-layer manner. For GCN, we use the batch training style modified from the original codes. For standard dataset, the sample size of baselines is as consistent as possible with the value suggested by the original paper. For GraphSAGE, we report the results by the Mean aggregator and the GCN aggregator with the sample sizes of $25$ and $10$ respectively for the first layer and second layer, following the default settings in \cite{hamilton2017inductive}. For FastGCN, the sample size of each layer in Cora, Citeseer, Pubmed and Reddit is 400, 400, 100, 400 respectively. For GraphSAGE and FastGCN, we test them with the original codes published by the authors. For our Ethereum dataset, we test the effectiveness of baselines with different sample size. For SGC, we conduct a comparative experiment similar to that set up in the original paper and use the batch training style same as GCN.

\subsubsection{Implementation details.}For all baselines and our method, the hidden dimension and learning rate for Reddit are set as 128 and 0.0001, and for the other three citation network and Ethereum datasets, they are 16 and 0.01. The batch size is 256 for standard datasets and 128 for Ethereum dataset. We use Adam~\cite{DBLP:journals/corr/KingmaB14} as the optimization method and no dropout is adopted. The weight decays for Ethereum dataset is set as $0.0005$. Activation function $\sigma$ is the LeakyReLU \cite{maas2013rectifier}.
Suggested by \cite{kipf2017semi}, we train all models using early stopping with a window size of 30. 
We run the experiments with different random seed over 20 times and record the averaged accuracy and standard variances. All the experiments are conducted on a machine with a NVIDIA Tesla P100 GPU SXM2 (16 GB memory), 8-core Intel(R) Xeon(R) CPU E5-2667 v4 (3.20GHz), and 252 GB of RAM.

\subsection{Comparisons}

%\paragraph{Datasets.} 
%		\begin{figure}
%		\centering
%		\includegraphics[width=0.5\textwidth]{fig/time_per_epoch.pdf}  
%		\caption{Training time per epoch on Pubmed and Reddit. One training epoch means a complete pass of all training samples}
%		\label{time_per_epoch}
%	\end{figure}
\subsubsection{Accuracy.}
\begin{table*}[tb]
	\centering
	\caption{Accuracy comparisons with state-of-the-art methods of four standard datasets. }
	\label{Accuracy for standard}
	\begin{tabular}{ccccc}  
		\toprule
		Methods & Cora & Citeseer & Pubmed&Reddit \\
		\midrule
		Batch GCN     & \textbf{86.8\%$\pm$0.34\%}  & 77.4\%$\pm$0.22\%  &86.5\%$\pm$0.33\% &  94.1\% $\pm$0.31\%  \\
		GraphSAGE-gcn  & 82.1\%$\pm$0.71\%  &75.1\%$\pm$0.56\%  & 88.2\%$\pm$0.24\% &  92.8\%$\pm$0.29\%  \\
		GraphSAGE-mean & 83.3\% $\pm$0.80\%  & 74.1\% $\pm$0.47\% & 88.6\% $\pm$0.33\% &  94.2\% $\pm$0.21\% \\
		FastGCN  & 83.3\%$\pm$0.47\% & 77.8\%$\pm$0.39\%  & 88.5\%$\pm$0.34\%  &  94.0\%$\pm$0.14\% \\
		SGC  & 81.0\%$\pm$0.32\% & 76.8\%$\pm$0.31\%  & 82.9\%$\pm$0.28\%  &  91.1\%$\pm$0.22\% \\
		\midrule
		MG-GCN  & 85.8\%$\pm$0.63\%& \textbf{78.2\%$\pm$0.41\%}  & \textbf{89.6\%$\pm$0.31\%}   &  \textbf{95.3\%$\pm$0.04\%}\\
		\bottomrule
	\end{tabular}
\end{table*}
Table \ref{Accuracy for standard} summarizes average value with standard variance of prediction accuracy in four standard datasets for the baselines and our proposed model. We observe that the proposed MG-GCN outperforms baselines for Citeseer, Pubmed and Reddit. For Cora, the accuracy of our model is slightly lower than Batch GCN, but still higher than other sampling-based algorithms and SGC. 

Compared with other sampling GCNs, the excellent results of our model should thanks in large part to the fine-aggregators, which aggregate all one-hop neighbors in the first layer. As discussed in Section 1, information from first-order neighbors is obviously more important than that of higher-orders. We observe that, for experiments on Cora, the accuracy of GraphSAGE and FastGCN is much lower than that of Batch GCN and MG-GCN, and this phenomenon may blame on the ignorance of first-order neighbors. 

However, for node classification on Pubmed and Citeseer, GraphSAGE and FastGCN can derive higher accuracy than Batch GCN, which is contrary to the results on Cora. A possible reason for this result is that information aggregated from the full neighbors in Batch GCN, especially that from the second-order or higher-order neighbor nodes, may include some noises which hinder the prediction performance. Recent researches~\cite{zugner2018adversarial,DBLP:conf/icml/DaiLTHWZS18} also suggested that the imperceptible noise could cause a great impact on GCN models. Different from Batch GCN, the proposed MG-GCN sample neighbors based on node degree, which can probably reduce the probability of sampling noisy neighborhood information to some extent. As discussed in Section~\ref{sec:model}, only a part of higher-order neighbors are sampled and aggregated, and the nodes with larger degrees, which are intuitively less likely to be noisy nodes, have a larger sampling probability. 

As for simplifying GCNs, SGC has the lowest accuracy on Cora, Pubmed and Reddit, which indicates that removing the nonlinear units between consecutive layers obviously decline the accuracy of this model.
\begin{table*}[tb]
	\centering
	\caption{Accuracy comparisons with state-of-the-art methods of Ethereum dataset.}
	\label{Accuracy for ether}
	\begin{tabular}{p{5cm}p{3cm}p{3cm}}  
		\toprule
		Methods & Sample Size & Accuracy  \\
		\midrule
		\multicolumn{1}{l}{Batch GCN}      & -& 70.4\%$\pm$0.04\% \\
		\midrule
		\multicolumn{1}{l}{SGC}      & -& 63.5\%$\pm$0.03\% \\
		\midrule
		\multicolumn{1}{l}{GraphSAGE-gcn}     & (5, 5)& 65.4\%$\pm$0.02\%\\
		& (25, 10)& 64.3\%$\pm$0.03\% \\
		& (25, 25)& 64.2\%$\pm$0.02\%\\
		& (100, 100)& 69.8\%$\pm$0.01\% \\  
		\midrule
		\multicolumn{1}{l}{GraphSAGE-mean }     & (5, 5)& 78.5\%$\pm$0.01\% \\
		& (25, 10)& 81.9\%$\pm$0.02\% \\
		& (25, 25)& 82.6\%$\pm$0.01\%\\
		& (100, 100)& 83.3\%$\pm$0.01\%\\
		\midrule
		\multicolumn{1}{l}{FastGCN}      &  (100, 100)& 54.1\%$\pm$0.08\% \\
		& (800, 800)& 69.9\%$\pm$0.05\%\\
		& (1600, 1600)& 74.8\%$\pm$0.01\% \\
		& (25600, 25600)& 77.3\%$\pm$0.02\% \\

		\midrule
		\multicolumn{1}{l}{MG-GCN}      & (5, -) & \textbf{85.4\%$\pm$0.02\%} \\
		\bottomrule
	\end{tabular}
\end{table*}

Table \ref{Accuracy for ether} demonstrates similar results for our method and baselines in Ethereum dataset. ($N_1$, $N_2$) means sampling $N_1$ nodes at first layer and $N_2$ at second layers. Note that Batch GCN and SGC does not have to sample nodes and MG-GCN just needs sample once for a 2-layer GCN model. To ensure a fair comparison, we test other methods with different sample sizes to obtain the best accuracy. The results show that our model can still derive better accuracy than other methods. In addition, we observe the poor results of FastGCN when the sample size is small and the accuracy increases obviously when sample size grows. We suppose that it may be caused by the layer independent sampling strategy, which makes it difficult to sample nearby nodes for training in a large dataset. Besides, we notice that the classification accuracy of GraphSAGE-mean is significantly better than GraphSAGE-gcn, which indicates  the superiority of Mean aggregator in large networks.

\subsubsection{Training and Convergence Speed.} 

\begin{table}[tb]
	\centering
	\caption{Training and convergence speed on Ethereum, Pubmed, and Reddit.}
	\label{speed and convergence}
	\begin{tabular}{p{2cm}p{3cm}p{3cm}p{3cm}}  
		\toprule
		Dataset & Method & Epoch Time(s) & Total Time(s)\\
		\midrule
		\multicolumn{1}{l}{Ethereum}&Batch GCN      & 0.47& 43.03\\
		&GraphSAGE  &0.21 &24.64  \\
		&FastGCN &0.32 & 49.80 \\
		&MG-GCN  &\textbf{0.16} &\textbf{14.20} \\
		\midrule
		\multicolumn{1}{l}{Pubmed} 	&Batch GCN     & 1.15 &44.72 \\
		&GraphSAGE  &1.07 &40.15  \\
		&FastGCN &0.40 &42.36  \\
		&MG-GCN  &\textbf{0.37} &\textbf{19.38} \\
		\midrule
		\multicolumn{1}{l}{Reddit}&Batch GCN      & 93.47& 5608.2\\
		&GraphSAGE  &10.51 &575.34  \\
		&FastGCN &\textbf{6.69} & 1138.33 \\
		&MG-GCN  &6.94 &\textbf{401.53} \\
		
		\bottomrule
	\end{tabular}
\end{table}
We test the training and convergence speed of baselines and our method by recording the training time per epoch and the total training time, respectively. Since the accuracy of SGC is much lower than other methods, we find it unfair to compare its speed and convergence time with other methods. Note that the results of GraphSAGE-mean and GraphSAGE-gcn are very close in terms of training and convergence speed, so we just record the result of GraphSAGE-mean to represent efficiency of GraphSAGE. 

Table \ref{speed and convergence} compares the training and convergence speed of different GCNs on three relatively large datasets, namely, Pubmed, Reddit and Ethereum. Training speed is evaluated by the training time per epoch. One training epoch means a complete pass of all training samples. The convergence speed is measured by the total training time. The sample size in Ethereum for GraphSAGE, FastGCN and MG-GCN is (100, 100), (25600, 25600) and 5 respectively. 

We observe that our model can achieve the comparable training speed, which is quantified by epoch time, with FastGCN while outperforming GraphSAGE and Batch GCN on Pubmed and Reddit. For Ethereum, the proposed model achieves faster training speed than all the baselines. In addition, FastGCN is beaten by GraphSAGE in terms of epoch time on Ethereum dataset because of its large sample size.

The high speed of our model benefits from the combination of coarse-grained aggregators and fine-grained aggregators. On the one hand, the coarse-grained aggregator can help us to focus on the sampled nodes as it works in other sampling-based GCN models. On the other hand, the fine-grained aggregator can implicitly reduce the sample size. Specifically, the sample size of our model is 6, 6, 4, 4 and 5 for Cora, Citeseer, Pubmed, Reddit, Ethereum respectively. The pretty small sample size of our model benefits from the fact that the first-order neighborhood information can be well reserved by the fine-grained.

The average time duration of each epoch is not enough to evaluate the efficiency of a method as it may be very slow or difficult to reach the convergence. In this work, the total training time given in Table \ref{speed and convergence} is utilized for comparing convergence.
We can observe from Table \ref{speed and convergence} that MG-GCN excels other GCN models in terms of the total training time, indicating that the proposed model has the fastest convergence speed.

Furthermore, although the training speed of GraphSAGE is relatively slow (indicated by the relatively long epoch time), it requires less time to converge than FastGCN on Pubmed and Reddit. 
As discussed above, the poor convergence performance of FastGCN may blame on its sampling strategy, which samples nodes from all nodes in a graph regardless of whether there is a neighbor relationship between them. As for GraphSAGE and our model, the sampling strategies are neighborhood dependent, which guarantees the connectivity.

\subsubsection{Memory Complexity Analysis.}
The memory complexity of a mini-batch GCN model consists of two parts, parameters of neural networks and embeddings of involved nodes in a batch. The former part is used to conduct the process of linear transformation and the latter part is employed to aggregate information in the forward pass and calculate gradients in the backward pass.

As for parameters of neural networks, all these models are the same in a 2-layer style. Therefore, the main differences of memory costs between GCN models depend on the spaces for storing embeddings of involved nodes in a batch. Furthermore, as the hidden dimensions of embeddings of all models are set the same, the memory complexity only relies on the number of involved nodes $N_{\rm in}$.

For the Batch GCN without any sampling method, the number of involved nodes $\left|N_{\rm in}\right|$ grow exponentially with the number of layers, and the memory complexity should be $\mathcal{O} (bd^L)$ in a batch, where $b$ is the batch size, $d$ is the average degree and $L$ is the number of layers. For SGC, it does not adopt any sampling strategy so SGC has the same memory complexity as Batch GCN. For GraphSAGE, the node-wise sampling strategy decreases the memory complexity to $\mathcal{O} (bs^L)$, where $s$ is the sample size in each layer. For FastGCN, the layer-wise sampling method avoids the exponential increment of $\left|N_{\rm in}\right|$ with the number of layers and the memory complexity should be $\mathcal{O} (sL)$.
\begin{figure}[tb]
	\centering
	\includegraphics[width=\textwidth]{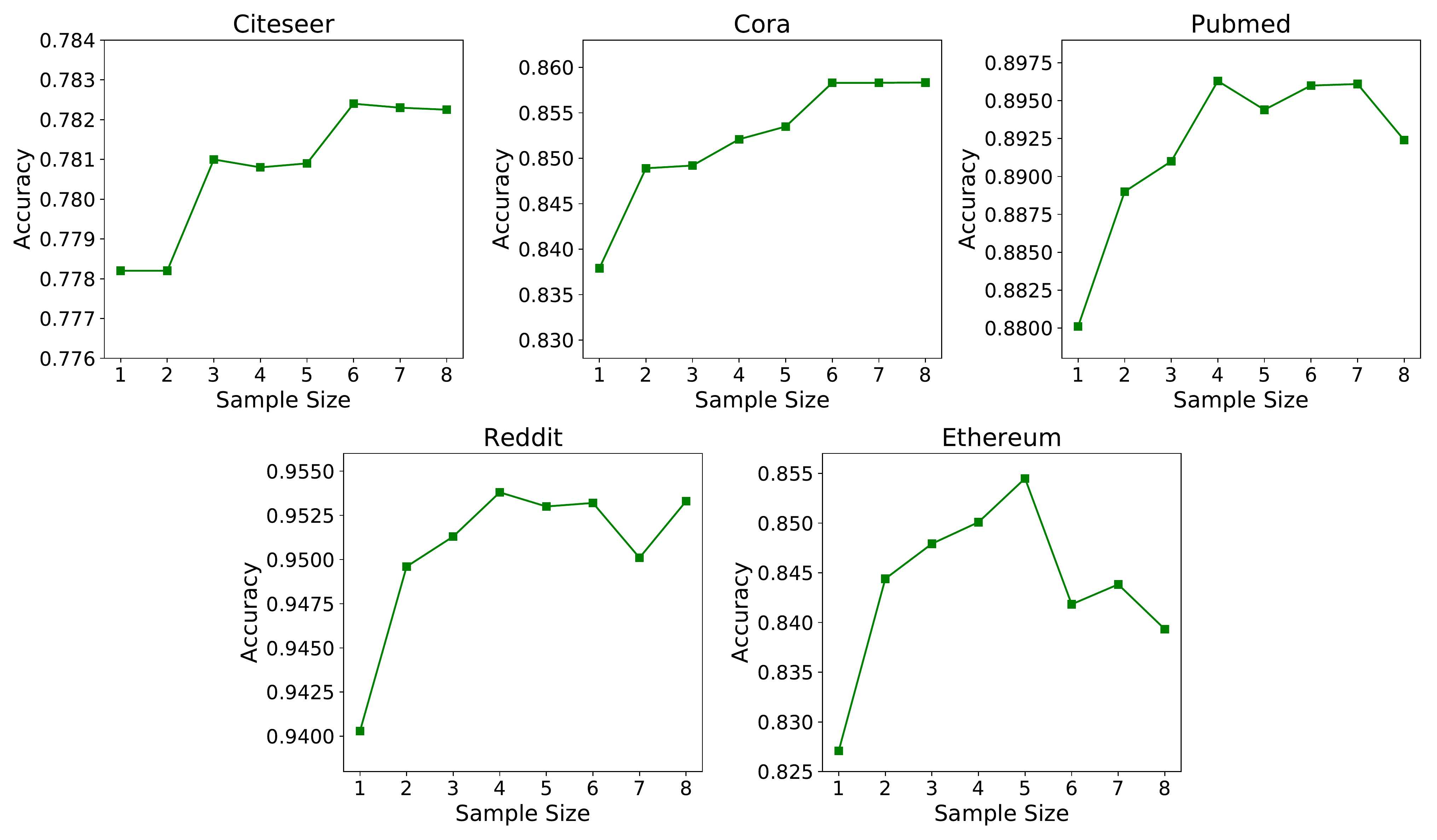}  
	\caption{Performance of MG-GCN with different sample size.}
	\label{Walks}
\end{figure}
As for our model, the number of involved nodes varies for different layers. In the first layer, where the fine-aggregator is applied, the number of involved nodes is $\mathcal{O} (bsd)$. However, for the nodes which are not sampled, we do not need to calculate their embeddings in the first layer because they will be abandoned in the following layers. Therefore, the memory cost in the first layer should be $\mathcal{O} (bs)$. For other layers which only focus on the sampled nodes, the number of involved nodes is $\mathcal{O} (bs)$. Consequently, the total memory complexity is $\mathcal{O} (bsL)$.

%The same as FastGCN, our method also avoids the exponential increment of involved nodes with the number of layers. 

\subsubsection{Effect of Sample Size.}

Next, we further test the effects of the sample size on the accuracy. As shown in Fig. \ref{Walks}, the accuracy of our model roughly increases with the sample size when the size is relatively small. When the sample size continues to increase, the accuracy starts to stop increasing, indicating that the sample size is enough.

\section{Conclusion}

%We present ClusterGCN, a new GCN training algorithm that is fast and memory efficient. Experimental results show that this method can train very deep GCN on large-scale graph, for instance on a graph with over 2 million nodes, the training time is less than an hour using around 2G memory and achieves accuracy of 90.41 (F1 score). Using the proposed approach, we are able to successfully train much deeper GCNs, which achieve state-of-the-art test F1 score on PPI and Reddit datasets.

In this paper, we propose MG-GCN, a novel GCN model with mix-grained aggregators for fast and effective learning on large graphs. Different from prior work which employs the identical aggregator for all GCN layers, we propose to use aggregators with different granularities for different layers. In addition, we present a new degree-based sampling strategy to collect useful neighboring nodes while avoiding the exponential growth of sampled nodes. Experimental results on four standard node classification benchmarks and a new blockchain transaction network demonstrated that, our model can achieve state-of-the-art performance in terms of accuracy, training and convergence, as well as memory complexity.  For future work, we plan to extend the mechanism of mixed-grained aggregators to other GCN models, trying to further improve the accuracy and convergence. 

%For future work, we consider two directions. On the one hand, the classification of different aggregators is generic. Therefore, we can extend this mechanism to other GCN models, such as GraphSAGE and FastGCN, trying to improve the accuracy and convergence. On the other hand, we are looking for more realistic and larger datasets to validate the effectiveness of our models.

\begin{acks}
The work described in this paper was supported by the National Key Research and Development Program (2016YFB1000101), the National Natural Science Foundation of China (61973325), the Key-Area Research and Development Program of Guangdong Province (No.2020B010165003) and the Fundamental Research Funds for the Central Universities under Grant No.17lgpy120.
\end{acks}

\bibliographystyle{ACM-Reference-Format}
\bibliography{sample-base}
\end{document}